\documentclass[sigconf,nonacm]{acmart}

\usepackage{booktabs}
\usepackage{graphicx}
\usepackage{orcidlink}
\usepackage{xurl}
\usepackage{morefloats}
\usepackage{dblfloatfix}

\renewcommand\footnotetextcopyrightpermission[1]{}
\begin{document}

\title[RibAssist 3D]{RibAssist 3D: Biplanar Rib-Fracture Detection, Addressing, and Selective 3D Localization from CT-Derived Projections}
\subtitle{A staged, sealed-cohort study of assistive selective 3D localization}

\author{Kabila Haile Soboka~\orcidlink{0009-0008-6740-3214}}
\orcid{0009-0008-6740-3214}
\affiliation{%
  \institution{The University of Texas at Austin}
  \city{Austin}
  \state{Texas}
  \country{USA}
}
\email{kabilahailesoboka@gmail.com}

\begin{abstract}
Rib fractures are common, clinically significant, and time-consuming to localize on computed tomography (CT).
This project asks a narrow, mechanistic question: can fractures detected independently in two orthogonal
projections (anteroposterior, AP, and lateral) be paired across views and triangulated into reliable 3D
fracture points at a controlled rate of false 3D outputs? We answer it with a staged diagnostic study rather
than a single end-to-end number. First, the biplanar geometry is exact and, given correct correspondence, highly
accurate: back-projection of correct paired centers gives 0.0~mm round-trip error, and detector-predicted centers
reconstruct to median 4.0~mm 3D error with 88\% within 10~mm and 93.6\% rib-exact. Second, on the sealed cohort a
substantial share of fractures is in principle recoverable: dual-view availability reaches 61.1\% and the
candidate graph contains a correct pair for 58.4\% of fractures. The binding limitation is therefore neither
geometry nor localization but confidence-limited cross-view correspondence: on the frozen detector, both
deterministic and learned pair-scoring reconstruct 0\% of fractures at a budget of one false 3D point per case. A
controlled factorial attributes the operational gain to lateral-detector quality rather than to the tested
deterministic or local-appearance matchers; retraining the lateral head lifts dual-view availability (0.52 to
0.76 in development) and moves the frontier from 0\% to 2.44\% recall at 10~mm, driven by deterministic
detector-confidence assignment. When the policy commits a correct pair, the emitted point is geometrically
accurate (sealed median 1.49~mm, rib-exact 93\%). As the terminal operating point of a deliberately conservative
fixed policy, a single pre-specified pass on the untouched 55-case cohort promotes 15 of 601 fractures to correct
3D localizations at 0.436 false 3D points per case: a 2.50\% end-to-end commitment yield (case-bootstrap 95\%
interval [0.69\%, 4.48\%], excludes zero), versus 0\% for the original detector's selected controlled-budget
policy. This end-to-end yield is the intersection of several selective gates, not a measure of the detector,
geometry, or rib addressing in isolation. Overall, the study establishes a confidence-gated framework for
selective 3D localization from CT-derived biplanar projections: reconstruction geometry is reliable, conditional
localization is accurate, and cross-view correspondence is the dominant operational bottleneck, with the current
implementation intended as an assistive workflow rather than a fully automated reconstructor.
\end{abstract}


\maketitle

\section{Introduction}

Rib fractures are among the most frequently encountered injuries in thoracic trauma, and increasing fracture
burden is associated with greater morbidity and mortality; fracture count and the presence of flail segments
therefore contribute to clinical risk assessment and management
decisions~\cite{sirmali2003ribfractures, ziegler1994ribfractures, flagel2005ribs}.
Yet localizing every fracture on a CT volume is tedious
and error-prone; subtle non-displaced fractures are easily missed on axial slices, and manually assigning each
finding to an anatomical rib level (``addressing'') is slow. Automated 3D localization would relieve this burden,
but full volumetric detection networks are computationally heavy and require dense 3D annotation.

An attractive alternative is a projection-based route: render the CT to two orthogonal orthographic views, detect
fractures independently in each, and triangulate the matched detections back into 3D. A projection-based approach
offers a lower-dimensional alternative that may reduce inference and annotation complexity relative to dense
volumetric detection. The route hinges on one hard sub-problem: solving the cross-view correspondence, i.e.,
deciding which AP detection and which lateral detection are the same physical fracture, at a controlled rate of
false 3D outputs. Because the deployed object is a 3D point, success must be measured operationally, never using
hidden ground-truth correspondence identity.

Crucially, the AP and lateral views used here are deterministic orthographic renderings of CT volumes, not
independently acquired clinical radiographs. The study therefore evaluates the reconstruction mechanism under
controlled geometry, not generalization to real biplanar radiography.

This paper reports a staged diagnostic study of exactly that sub-problem, built on these simulated biplanar
projections derived from the public RibFrac and RibSeg datasets. Rather than claiming a deployable detector, we
treat the system as an object of investigation and ask precisely where a biplanar 3D rib-fracture reconstructor
succeeds and where it fails. Our contributions are: (1) a decomposition of the pipeline into geometry,
localization, and correspondence, with a controlled experiment isolating each; (2) evidence that the operational
gain is attributable to lateral-detector quality rather than to the tested deterministic or local-appearance
correspondence methods; (3) a lateral retraining intervention that produces the first nonzero controlled-budget
reconstructions; and (4) a single pre-specified confirmation on an untouched sealed cohort, with fail-closed
provenance, that reproduces the development estimate out of sample.

RibAssist~3D is evaluated primarily as a confidence-gated 3D reconstruction system, but its utility is broader than
automatic triangulation: even when correspondence cannot be committed, the per-view detections still highlight
suspicious regions and support rib-level addressing. The workflow therefore treats 3D localization as an additive
output rather than a prerequisite, so the system's value does not collapse to zero when it abstains.

\section{Related Work}

\textbf{Deep rib-fracture detection on CT.} The closest reference point is FracNet~\cite{jin2020fracnet}, which
established the RibFrac benchmark and a 3D U-Net-style detector that segments and detects rib fractures directly in
the CT volume. FracNet operates fully in 3D and reports free-response operating characteristic (FROC) sensitivity
against radiologist labels. Our work shares the detection target and the FROC-style, budget-aware evaluation
philosophy, but deliberately departs from volumetric inference: we detect in 2D projections and study whether the
cheaper biplanar route can recover 3D points at a controlled false rate. FracNet also motivates our operational
endpoint, since a clinical 3D CAD system is judged by how many true fractures it localizes per allowed false
finding, not by pixel overlap.

\textbf{Rib labeling and anatomical centerlines.} Assigning a fracture to a specific rib requires rib identity.
RibSeg v2~\cite{jin2023ribsegv2} provides large-scale rib labels and anatomical centerlines and benchmarks
point-cloud methods for rib segmentation and centerline extraction. We use RibSeg-derived rib geometry both as the
substrate for our ``addressing'' model, which predicts side and rib level for each detection, and as the anatomical
ground truth against which reconstructed points are checked for rib-exactness. Where RibSeg v2 treats rib parsing
as the end goal, we consume it as a supporting signal for fracture localization and as an evaluation oracle.

\textbf{Biplanar-to-3D reconstruction.} The idea of recovering 3D structure from two orthogonal views is embodied
by X2CT-GAN~\cite{ying2019x2ctgan}, which reconstructs an entire CT volume from biplanar X-rays with a generative
adversarial network. X2CT-GAN is generative and dense: it hallucinates a full volume, optimizing image-similarity
losses. Our problem is discriminative and sparse: we do not reconstruct anatomy, we triangulate a small set of
fracture points and must decide correspondence explicitly. This distinction matters for evaluation, since a
generative volume can look plausible while a triangulated point is either correct within tolerance or not. Our
study can be read as isolating the correspondence bottleneck that a sparse biplanar reconstructor faces, a
bottleneck a generative model hides inside its learned prior.

\textbf{Cross-view correspondence and selective prediction.} Matching observations across views is a classical
geometry problem: with calibrated cameras, epipolar constraints reduce the search for a point's partner to a line,
and one-to-one assignment then resolves the match~\cite{hartley2004multiview}. Our orthographic setup gives an
even simpler constraint, agreement on the shared superior-inferior axis, but as we show that constraint is far
weaker than a full epipolar geometry because many distinct fractures share the same axis coordinate. The second
relevant thread is selective prediction, where a model may abstain rather than emit a low-confidence output to
trade coverage for reliability~\cite{geifman2017selective}. Our assignment-with-abstention is exactly such a
mechanism at the level of correspondence commitments, and the controlled false-output budget is the coverage
constraint under which we evaluate it.

\section{Methodology}

\subsection{Data and projection}
Each CT volume from RibFrac, together with its fracture annotations and RibSeg-derived rib segmentation, is
rendered to AP and lateral orthographic projections at $256\times256$. The superior-inferior (SI) axis is shared
between the two views, which is the geometric key that later gates candidate pairs.

Table~\ref{tab:splits} summarizes the case-disjoint splits. Splits are patient-disjoint at the CT level: every
volume, and therefore every projection, peak, and crop derived from it, belongs to exactly one split, so no
same-scan material crosses the development-test boundary. The detector is trained on the development pool; all
staged diagnostics, calibration, and policy or model selection are performed out of fold on the held-out
diagnostic split (the L2 lateral head is retrained on the development-internal training slice and never sees the
diagnostic or sealed cases); and the addressing model is trained on the same development pool. The sealed cohort is
opened once, for a single fixed-policy pass. We note explicitly that the sealed cohort contains no
fracture-negative studies, so per-case false-output behavior on truly negative scans is not measured.

\begin{table}[t]
  \caption{Case-disjoint data splits. The diagnostic split is a held-out slice of the development pool used for all
  out-of-fold selection; the sealed cohort is opened once. Negative denotes fracture-free cases.}
  \label{tab:splits}
  \small
  \setlength{\tabcolsep}{3pt}
  \begin{tabular}{lcccl}
    \toprule
    Split & Cases & Fract. & Neg. & Purpose \\
    \midrule
    Training pool & 325 & 2,686 & 20 & model training \\
    Diagnostic (held out) & 65 & 492 & 4 & OOF selection \\
    Sealed test & 55 & 601 & 0 & confirmation \\
    \bottomrule
  \end{tabular}
\end{table}

\subsection{Pipeline}
\begin{figure*}[t]
  \centering
  \includegraphics[width=0.98\textwidth]{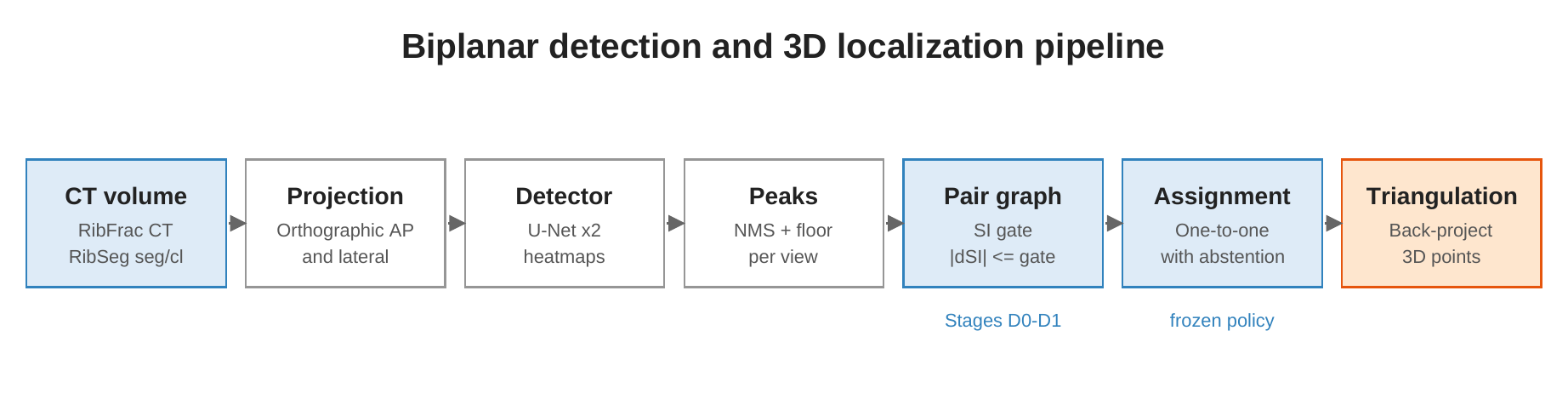}
  \caption{End-to-end pipeline. Two orthographic projections are detected independently, peaks are paired by
  shared-axis (SI) geometry into a candidate graph, and an assignment with abstention commits only confident pairs
  for triangulation. The operational bottleneck is that too few true cross-view edges receive enough detector
  confidence to clear the commit threshold.}
  \label{fig:arch}
\end{figure*}

Figure~\ref{fig:arch} shows the workflow. A per-view U-Net~\cite{ronneberger2015unet} detector emits a fracture
heatmap for each view; local peaks are extracted with per-view non-maximum suppression (NMS) and a score floor.
Peaks are paired into a candidate graph gated by SI agreement ($|\Delta \mathrm{SI}|$ below a tolerance). A
one-to-one \emph{assignment with abstention} commits a correspondence only when its matching cost clears a
threshold, so low-confidence pairs produce no 3D output. Each committed pair is back-projected and triangulated to
a 3D point. A separate addressing model predicts the rib level of each detection. The detector, extraction policy,
correspondence configuration, and evaluation are all frozen before the sealed test, and every stage verifies data,
checkpoint, and protocol hashes before producing any number (fail-closed provenance).

\begin{figure*}[t]
  \centering
  \includegraphics[width=\textwidth]{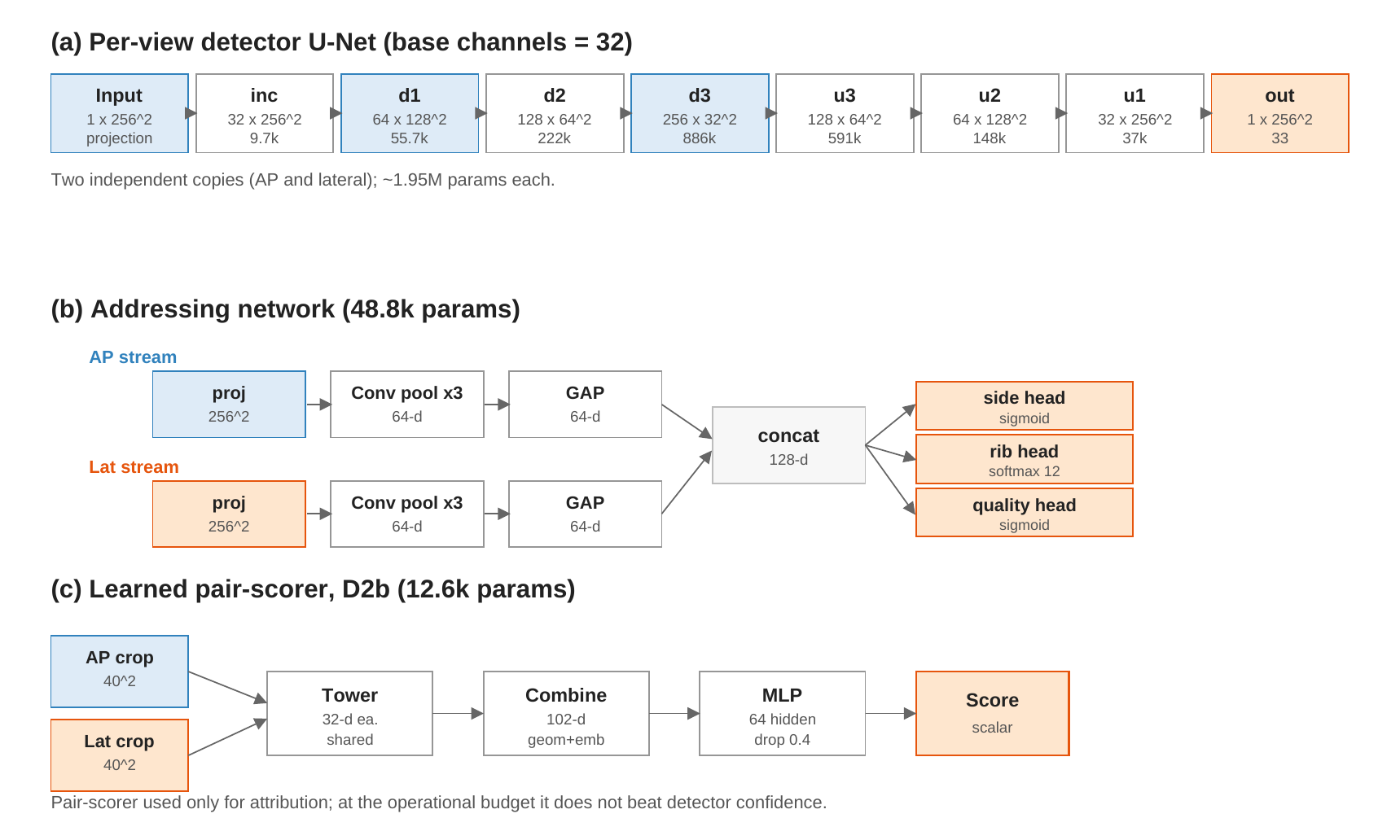}
  \caption{Layer-by-layer architecture of the three trained networks. The biplanar detector is two independent
  single-channel U-Nets (base channels 32, about 1.95M parameters each); ``fusion'' is a geometric candidate union
  on the SI axis, not a learned layer. The addressing network (48.8k parameters) and the learned pair-scorer
  (12.6k parameters) are the two auxiliary models.}
  \label{fig:layers}
\end{figure*}

\subsection{Networks}
The detector (Figure~\ref{fig:layers}) is a from-scratch U-Net with base channel width 32, deployed as two
independent copies (AP and lateral), about 1.95M parameters per view. Each stage is a double convolution
(Conv~$3\times3$, BatchNorm, ReLU, twice); the encoder halves resolution with max-pooling and the decoder restores
it with bilinear upsampling and skip concatenation, ending in a $1\times1$ convolution and sigmoid. Training uses a
penalty-reduced focal loss~\cite{lin2020focal}. The addressing network is a compact dual-stream CNN (about 48.8k
parameters) whose two 64-dimensional view embeddings concatenate into three linear heads (side, rib level, and a
quality score); it is included as a supporting workflow component rather than a primary research
contribution. Because no independent sealed-cohort evaluation of the addressing head was performed, we
intentionally do not report standalone addressing accuracy; instead, its contribution is evaluated through the
complete end-to-end assistive workflow. The learned pair-scorer, used
only to test whether appearance beats geometry for correspondence, is a shared two-layer tower embedding
$40\times40$ crops with a small multilayer-perceptron head (about 12.6k parameters).

\subsection{Operational endpoint}
From Stage~D1 onward we use a single frozen endpoint. For the accepted 3D points in a case, we back-project each
accepted AP-lateral pair to a world-space point, compute its distance to every ground-truth (GT) fracture volume
(nearest voxel via a per-fracture KD-tree), and run an independent one-to-one prediction-to-GT matching at 5, 10,
and 15~mm. Recall at $X$ is distinct GT matched within $X$~mm divided by all GT; a false 3D point is an accepted
prediction unmatched at the tolerance. The primary endpoint and false-output budget are at 10~mm and at most one
false 3D point per case. This budget is a pre-specified, stringent system-development criterion, not a clinically
validated tolerance; it encodes the design goal that a practical reconstructor should rarely place a spurious 3D
point, and the analyses below are reported against it rather than against a clinically established threshold.

\subsection{Staged diagnostic design}
\begin{figure*}[t]
  \centering
  \includegraphics[width=0.86\textwidth]{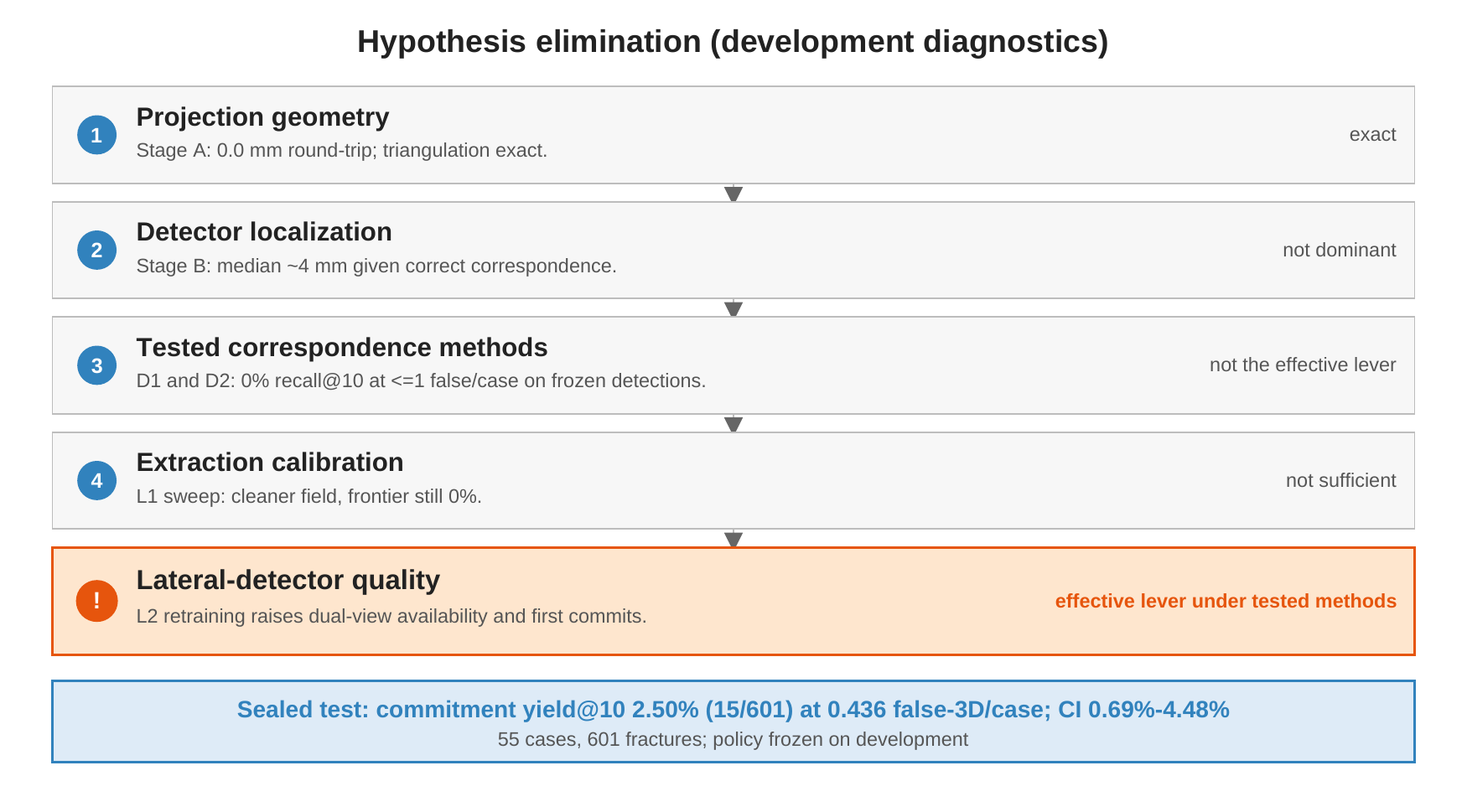}
  \caption{Identifying the effective operational bottleneck through staged elimination. Geometry is exact and localization is not the dominant
  barrier; the tested deterministic and local-appearance correspondence methods are not the effective lever, and
  candidate-field density alone is not sufficient; lateral-detector quality is what moves the frontier.}
  \label{fig:bottleneck}
\end{figure*}

Rather than tuning end-to-end, we run a diagnostic elimination (Figure~\ref{fig:bottleneck}) that swaps one
component at a time from oracle to model, so each experiment charges the failure to a single stage: geometry
(Stage~A), localization under oracle correspondence (Stage~B), the deployed correspondence (Stage~C), the
candidate graph (Stage~D0), deterministic and learned pair-scoring (Stages~D1--D2), lateral calibration
(Stages~L0--L1), and finally the lateral retraining intervention and a detector-by-correspondence factorial
(Stages~L2--L3).

\subsection{Experimental setup and reproducibility}
Both detectors are from-scratch U-Nets (base width 32, BatchNorm) trained with a penalty-reduced focal
loss~\cite{lin2020focal} and Adam. The Stage~L2 lateral head is warm-started from the frozen lateral weights and
trained for 60 epochs with a positive-branch weight of 3.0 and hard-negative emphasis re-mined every ten epochs on
the false peaks that survive the extraction policy; the final epoch is taken without validation-based checkpoint
selection. Peaks are extracted with AP non-maximum suppression radius 5 and floor 0.05, and lateral radius 3 and
floor 0.10 (the standing L2 policy). The assignment introduces a dummy unmatched node at per-node cost $u$ for
every real node, so a real edge of cost $c$ is preferred only when $c < 2u$; equivalently, a correspondence
commits only when its geometric-mean confidence clears $1-2u$ (0.167 at the operating point). We sweep $u$ and the
SI gate and select the case-level configuration out of fold on the operational objective (false 3D points at
10~mm), never on the sealed data. Every stage recomputes and checks the SHA-256 of its inputs (dataset, detector
checkpoints, frozen policy files) before emitting any number and aborts on mismatch; the sealed evaluation
additionally binds to an external data anchor so the test cohort cannot be silently swapped. This fail-closed
provenance makes the single confirmatory pass auditable. Confidence intervals are case-level bootstraps with 1{,}000
resamples over cases. Full training schedules, fold construction, heatmap-target generation, and the frozen policy
files are recorded in the project repository so every reported number can be regenerated.

\section{Results}

\subsection{Geometry is exact; localization is not the dominant barrier}
Pure orthographic back-projection scored against GT geometry gives a 0.0~mm transform round-trip and, from correct
paired GT centers, median 0.0~mm 3D error with 98.6\% rib-exact: triangulation is sound (Stage~A). Replacing oracle
centers with detector peaks while keeping oracle correspondence gives median 4.0~mm, 88\% within 10~mm, rib-exact
93.6\%, and rib$\pm1$ 100\% (Stage~B). Localization error is therefore real but tolerable relative to the
correspondence failure that follows, with the lateral view the weaker localizer; it is not the dominant barrier,
not a claim that localization is perfect.

\subsection{Correspondence on the frozen detector fails}
The deployed SI-only fusion matcher correctly and uniquely pairs only 4 of 205 dual-view fractures, and end-to-end
within-10~mm is about 0.6\% (Stage~C); when a pair is correct the reconstruction is still good, confirming the
failure is in getting correct pairs, not triangulation. The frozen detector emits about 40 AP but about 188
lateral peaks per case, so the SI-gated graph is dominated by non-fracture pairs, and $|\Delta\mathrm{SI}|$ cannot
separate same-fracture from cross-fracture pairs (identical medians). Dual-view availability is only 51.8\%, an
upper bound on recall for the current candidate-generation pipeline (Stage~D0). One-to-one assignment with abstention reconstructs 0\% at the budget, and the learned
two-tower appearance scorer is essentially at chance on the hard positive-versus-cross task (AUROC 0.57), so the
tested pair-scoring family is operationally exhausted on the frozen detections (Stages~D1--D2). Table~\ref{tab:d1} shows the shape of the failure: there
is no gentle operating point. Recall is exactly zero at any tolerable budget and only becomes nonzero once dozens
of false points per case are accepted, and even fully unconstrained the min-cost pairing realizes only 8.3\% of an
already-modest candidate ceiling, meaning it selects the wrong partner for most fractures.

\begin{table}[t]
  \caption{Deterministic correspondence (Stage~D1) operating frontier, out of fold on the development split. The
  transition from ``accept nothing'' to ``many false points'' is abrupt; there is no controlled-budget operating
  point.}
  \label{tab:d1}
  \begin{tabular}{lcc}
    \toprule
    False 3D points / case & Recall @10~mm & Fractures matched \\
    \midrule
    $\leq 3$ & \textbf{0.0\%} & 0 \\
    5 & 2.4\% & 12 \\
    10 & 6.3\% & 31 \\
    $\infty$ (accept all) & 8.3\% & 41 \\
    \bottomrule
  \end{tabular}
\end{table}

\subsection{The flood is a calibration artifact, but recalibration is not enough}
The lateral heatmap's per-case maximum sits near 0.091 (AP near 0.176); its output is compressed just above the
0.05 extraction floor, and 68\% of spurious peaks live in the single bin [0.05, 0.055) (Stages~L0--L0.1).
Re-tuning the lateral extraction policy (NMS radius 5 to 3, floor 0.05 to 0.10) shrinks the candidate field about
fourfold and recovers merged-away availability, yet the operational frontier stays at 0\% and is floor-invariant
(Stage~L1). Table~\ref{tab:l1} makes the dissociation explicit: across policies the candidate field collapses and
dual-view availability rises, while controlled-budget recall does not move off zero. Candidate-field density alone
is therefore not the binding lever under the tested policies: what limits the budget is how many true edges the
detector is confident enough to commit.

\begin{table}[t]
  \caption{Lateral extraction-policy sweep (Stage~L1), AP fixed at the deployed policy. Recalibration shrinks the
  candidate field and lifts availability, but controlled-budget recall stays at zero for every policy. R@10$_{\leq1}$
  is recall at 10~mm at one false 3D point per case; R@10$_{\infty}$ is uncapped.}
  \label{tab:l1}
  \small
  \setlength{\tabcolsep}{4.5pt}
  \begin{tabular}{lcccc}
    \toprule
    Policy & Pk/case & Dual-view & R@10$_{\leq1}$ & R@10$_{\infty}$ \\
    \midrule
    nms5 / 0.05 (deployed) & 187.6 & 0.518 & \textbf{0.0\%} & 8.3\% \\
    nms5 / 0.060 & 32.0 & 0.492 & \textbf{0.0\%} & 8.1\% \\
    nms5 / 0.065 & 20.1 & 0.468 & \textbf{0.0\%} & 7.3\% \\
    nms3 / 0.060 & 44.2 & \textbf{0.589} & \textbf{0.0\%} & \textbf{12.2\%} \\
    \bottomrule
  \end{tabular}
\end{table}

\subsection{The detector intervention is the effective lever under the tested methods}
\begin{figure*}[t]
  \centering
  \includegraphics[width=0.68\textwidth]{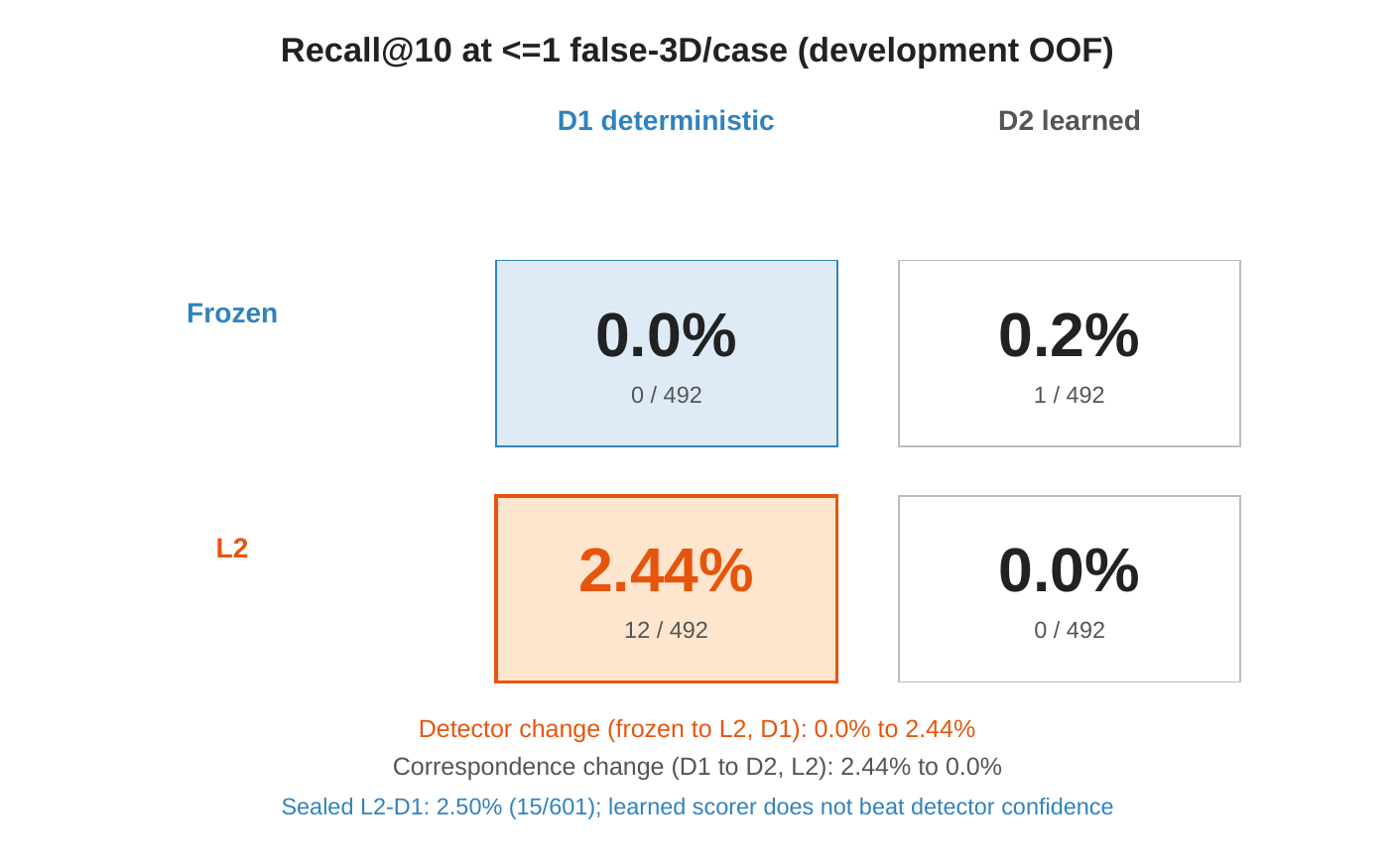}
  \caption{Detector-by-correspondence factorial. Only the retrained (L2) detector paired with deterministic
  detector-confidence assignment moves off zero; the learned pair-scorer does not beat detector confidence at the
  operational budget.}
  \label{fig:2x2}
\end{figure*}

Retraining only the lateral head with hard-negative mining on retained false peaks plus positive-branch
strengthening lifts the amplitude ceiling (0.091 to 0.155, on par with AP) and dual-view availability (0.52 to
0.76). For the first time the deterministic assignment commits where the frozen head could only abstain, moving
the development operational frontier from 0\% to 2.44\% recall at 10~mm at the one-false-per-case budget
(Stage~L2). The detector-by-correspondence factorial (Figure~\ref{fig:2x2}) attributes the observed operational gain to the detector intervention within the tested $2\times2$ design: the
only cell that moves off zero is the retrained detector with deterministic assignment (Table~\ref{tab:l3}).
Learned appearance does not beat detector confidence at the budget; its positive-versus-cross AUROC rises with
detector quality (0.57, 0.66, 0.71) but never enough to commit, and on the retrained head the learned scorer is
actually worse operationally than the detector's own geometric-mean confidence (Stage~L3). The floor-invariance of
the L2 operating point (the same 12 fractures commit across floors 0.06 to 0.12) confirms that the budget is set by
how many edges clear the confidence threshold, not by competitor density.

\begin{table}[t]
  \caption{Detector-by-correspondence factorial (Stage~L3): recall at 10~mm out of fold at the one-false-per-case
  budget, each head at its own calibrated policy. Only the retrained (L2) detector with deterministic assignment
  moves off zero (uncapped recall in parentheses).}
  \label{tab:l3}
  \begin{tabular}{lcc}
    \toprule
    Lateral head & Deterministic (D1) & Learned (D2) \\
    \midrule
    Frozen & 0.0\% (12.2\%) & 0.2\% (17.3\%) \\
    L2 retrained & \textbf{2.44\%} (21.5\%) & 0.0\% (20.7\%) \\
    \bottomrule
  \end{tabular}
\end{table}

\subsection{Sealed-test confirmation}
\begin{figure*}[t]
  \centering
  \includegraphics[width=0.95\textwidth]{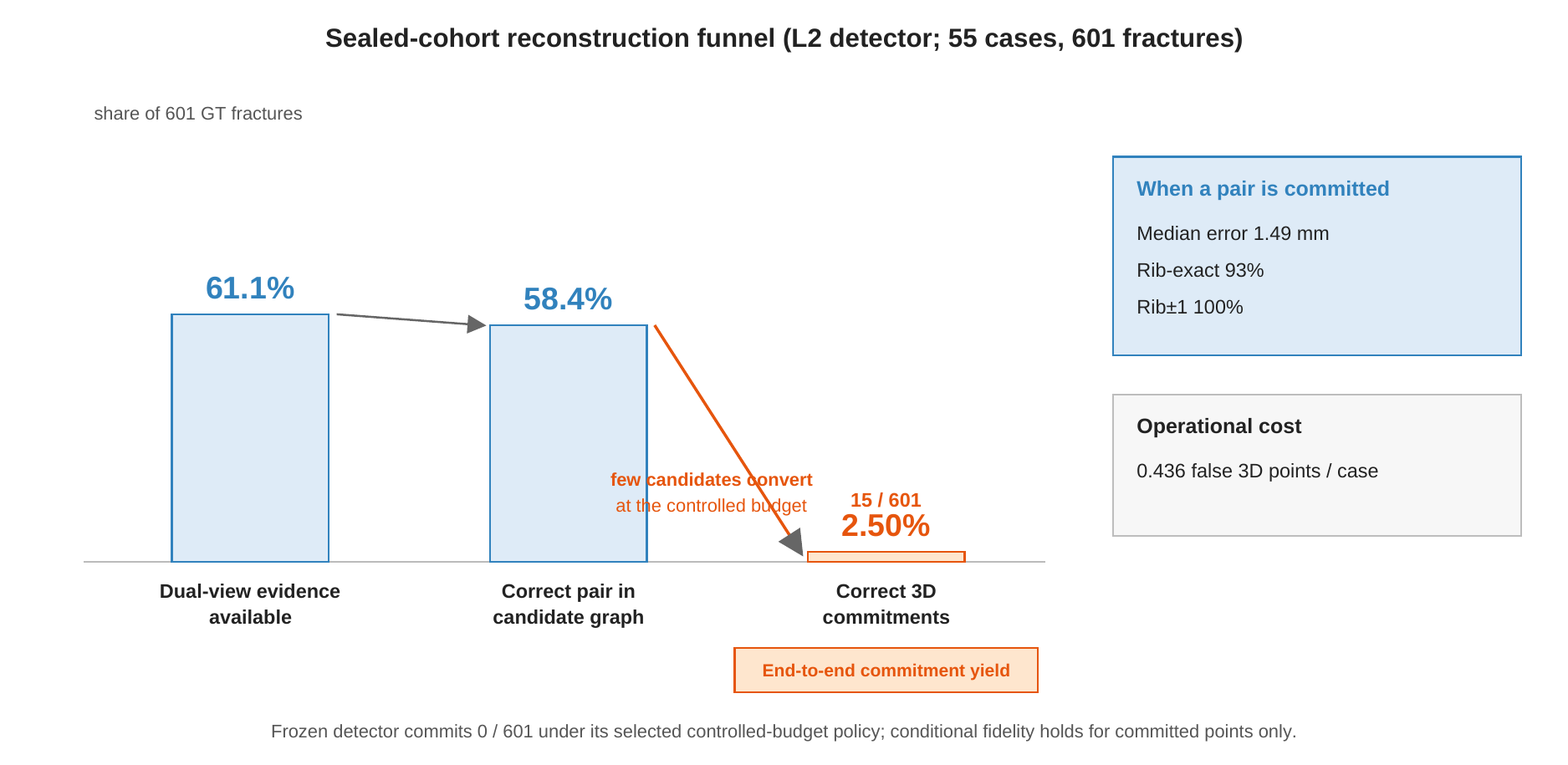}
  \caption{Sealed-cohort reconstruction funnel for the retrained (L2) policy. A large share of fractures reaches a
  recoverable candidate state (61.1\% dual-view available, 58.4\% with a correct pair in the candidate graph), but
  the controlled-budget assignment converts few of these into correct 3D commitments: the end-to-end commitment
  yield is 2.50\% (15 of 601). Committed points are accurate (median 1.49~mm, 93\% rib-exact) at 0.436 false 3D
  points per case; the frozen policy commits 0 of 601.}
  \label{fig:main}
\end{figure*}

A single pre-specified pass on the untouched 55-case cohort, with the extraction policy and correspondence
configuration frozen on development and no test-set reselection, reproduces the result
(Figure~\ref{fig:main} and Table~\ref{tab:sealed}). We read it in pipeline order rather than end-to-end first.
\emph{Evidence availability is substantial:} the retrained (L2) detector reaches 61.1\% dual-view availability
and a 58.4\% candidate ceiling at 10~mm, so most fractures are in principle recoverable from the candidate space.
\emph{Conditional localization is accurate:} when the policy commits a correct pair, median error is 1.49~mm with
rib-exact 0.93 and rib$\pm1$ 1.00. \emph{The end-to-end yield is the terminal metric of this funnel:} under the
deliberately conservative fixed policy the L2 detector promotes 15 of 601 fractures to correct 3D localizations, a
2.50\% correct 3D commitment yield at 10~mm at 0.436 false 3D points per case, with a case-bootstrap 95\% interval [0.69\%, 4.48\%]
above zero, whereas the frozen selected policy emits no predictions at all (0 of 601, a degenerate zero interval).
That 2.50\% is the joint outcome of several selective gates, i.e., detection in both views, inclusion of the true
pair in the candidate graph, survival of one-to-one assignment, confidence above the abstention threshold, and
localization within tolerance; it is a conservative end-to-end commitment yield, not a measurement of the
detector, geometry, or rib addressing in isolation. The result supports nonzero performance under the fixed L2
policy; it is not a paired hypothesis test of the detector difference, though the per-case direction is one-sided
(6 cases improve, none worsen). The sealed point estimate was similar to the development out-of-fold estimate
(2.50\% versus 2.44\%), and the detector-side mechanism transferred intact.

\begin{table}[t]
  \caption{Sealed cohort (55 cases, 601 GT fractures, no fracture-negative cases). Frozen detector's selected controlled-budget
  policy versus the retrained (L2) detector, under the pre-specified protocol (fixed policy; no test reselection). The
  commitment yield @10~mm is end-to-end fracture recall under the committed-output policy, with an all-fracture
  denominator (all 601 GT fractures).}
  \label{tab:sealed}
  \setlength{\tabcolsep}{3pt}
  \begin{tabular}{lcc}
    \toprule
    Metric & Frozen detector & L2 detector \\
    \midrule
    \shortstack[l]{Correct 3D commitment\\ yield @10~mm} & 0.0\% (0) & \textbf{2.50\%} (15) \\
    Commitment-yield 95\% CI & [0.0, 0.0] & \textbf{[0.69, 4.48]} \\
    False 3D points / case & 0.0 & 0.436 \\
    Cases with $\geq 1$ correct & 0 / 55 & 6 / 55 \\
    Matched median distance & -- & 1.49~mm \\
    Matched rib-exact & -- & 0.93 \\
    Dual-view availability & 0.542 & 0.611 \\
    Candidate ceiling @10 & 0.265 & 0.584 \\
    \bottomrule
  \end{tabular}
\end{table}

\subsection{Impact}
The primary scientific value of this study is diagnostic, while its application-level value is an
assistive workflow rather than a deployable clinical system. It converts a vague ``biplanar 3D does not
work'' into a specific, reproducible claim: geometry and triangulation are exact, localization under correct correspondence is spatially accurate and rib-exact, and
the operational failure manifests at the cross-view correspondence stage, so the gain came from lateral-detector
quality rather than from the tested matching methods.

At the application level, the system demonstrates a layered assistance model: per-view detections provide
highlighting, rib addressing supplies anatomical organization, and only sufficiently confident AP-lateral pairs are
promoted to 3D points. The low end-to-end commitment yield should not be read as an indiscriminate failure to
localize; it is a consequence of the intended operating policy, an explicit safety mechanism rather than an
implementation artifact. The system deliberately abstains from 3D
localization when cross-view evidence is insufficient, favoring omission over unsupported geometric reconstruction,
so every committed point clears a substantially higher evidential standard than a policy requiring complete coverage
(sealed committed points had median error 1.49~mm and 93\% rib-exactness). This selective-prediction (reject-option)
design keeps lower-level 2D findings available for conventional review when the system abstains rather than emitting
potentially misleading geometry, and supports \emph{selective} localization rather than comprehensive
reconstruction. These are demonstrated properties of the workflow's design, not claims of measured clinical benefit.

\subsection{Clinical workflow relevance}
RibAssist~3D is intended as a secondary-review and anatomical-organization tool for chest-trauma CT, not an
autonomous diagnostic system. A potential user is a radiologist, emergency physician, or trauma clinician reviewing
suspected thoracic injury: the CT is read conventionally, and the system runs alongside it to highlight suspected
fractures, predict side and rib level, and place selective 3D locations, while findings without confident
correspondence stay visible rather than being discarded. Plausible applications are secondary fracture review,
rib-level documentation, structured injury summaries, and prioritization for specialist review where coverage is
limited. These are hypothetical here: with no clinician reader or workflow study, we claim workflow
\emph{relevance}, not demonstrated clinical benefit.

\subsection{Clinician-review prototype}
\begin{figure}[t]
  \centering
  \includegraphics[width=\columnwidth]{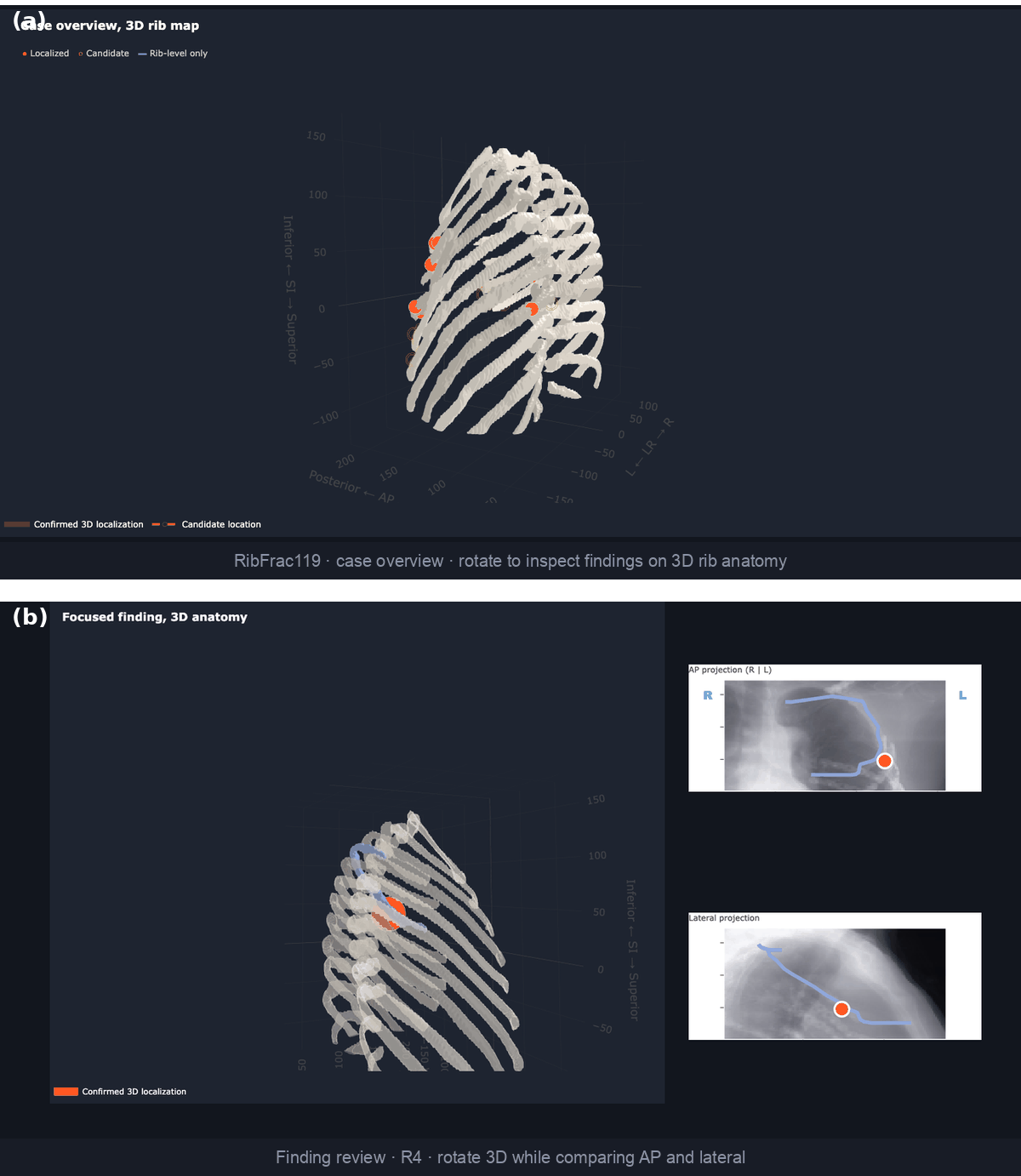}
  \caption{Clinician-review prototype (live model outputs). (a) Case overview on 3D rib anatomy, with
  model-localized, candidate, and rib-level-only findings shown as distinct states. (b) Per-finding review: the AP
  and lateral viewers with markers beside the 3D anatomy and the committed (localized) 3D point. Detection and
  addressing remain primary; 3D localization is additive and abstains without dropping a detection.}
  \label{fig:review}
\end{figure}

To situate the mechanism in a workflow, we built an interactive review interface (Figure~\ref{fig:review}) that
runs the trained models live. Detection and rib addressing remain the primary output; biplanar 3D localization is
an \emph{additive} layer. A detection is never removed because the 3D correspondence abstains: findings are shown
in three explicit states, model-localized (a committed, triangulated 3D point), candidate (a cross-view pair that
did not clear the commit threshold), and rib-level-only (a 2D detection with addressing but no 3D), so an abstained
pair still surfaces as a reviewable finding rather than disappearing. The interface presents the AP and lateral
viewers with their markers, the 3D rib anatomy with committed and candidate locations, the detection and
addressing confidences, cross-view status, and accept, needs-review, and reject actions, along with provenance and
audit information. It is a workflow demonstration, and no clinical decisions were made with it.

\section{Limitations and Threats to Validity}
Several constraints bound the strength of the claim. \emph{Simulated projections:} the AP and lateral views are
orthographic renderings of CT, not real biplanar radiographs, so scatter, tissue overlap, and calibration error of
a real acquisition are not modeled. \emph{Small sealed cohort:} at $n=55$ the confidence interval is wide and the
estimate is directional rather than precise. \emph{Negative-scan safety unvalidated:} the sealed cohort contains no
confirmed fracture-negative studies, so the per-case false-output behavior on truly negative scans is unmeasured, a
gap that matters for any triage use. \emph{Development selection:} the retrained head was chosen on the development
split, and the sealed pass confirms the direction of the effect but not an unbiased effect size. \emph{Joint
mediation:} retraining changed availability, prevalence, confidence, and the candidate ceiling together, so the
gain is attributed to detector quality overall, not to availability as an isolated causal mediator. \emph{Metric
scope:} recall at 10~mm with independent prediction-to-ground-truth matching rewards localization within tolerance
at a controlled false rate, but does not assess fracture typing, displacement, or acute-versus-healed status.

\section{Ethics and Intended Use}
This retrospective study uses public, de-identified research datasets (RibFrac and RibSeg v2) under their
respective licenses, with no patient interaction and no prospective or patient-facing decisions. RibAssist~3D is a
research prototype, not a medical device, and must not be used for diagnosis or patient care. The
clinician-review interface demonstrates human-in-the-loop review but has not undergone usability or clinical
validation. Because the AP and lateral inputs are simulated CT-derived projections, no claims are made about
performance on independently acquired radiographs; no demographic subgroup or fairness analysis was performed
because the available dataset metadata and evaluation did not support one, which is itself a limitation for
equitable deployment.

\section{Conclusion and Future Work}

We presented RibAssist~3D, a staged diagnostic study of biplanar 3D rib-fracture reconstruction from CT-derived
projections. By swapping components from oracle to model one at a time, we identified the effective operational
bottleneck under the tested detector and correspondence methods:
projection geometry is exact and detector localization is rib-accurate, so the operational failure manifests at
the cross-view correspondence stage, where weak lateral detections limit confident commitment, and the gain came
from lateral-detector quality rather than from the tested deterministic or local-appearance correspondence methods. A lateral retraining intervention produced the first nonzero
controlled-budget reconstructions, moving the frontier from 0\% to 2.44\% in development, and a single
pre-specified evaluation on an untouched 55-case cohort confirmed a 2.50\% correct 3D commitment yield at 10~mm at 0.436 false 3D points
per case, with a bootstrap interval above zero. We additionally implemented an interactive review prototype that
preserves all original detections, shows model-localized and uncommitted candidate locations as distinct states,
and supports human review without presenting the system as diagnostically validated. The sealed result therefore
establishes feasibility for confidence-gated, selective 3D localization and for an assistive workflow, highlighting
suspected findings in each view, assigning anatomical rib information, preserving uncertain detections for review,
and adding accurate 3D localization only when confidence is sufficient, while identifying the detector and
global-correspondence improvements required for broader coverage.

Within the scope tested, the study delivers concrete empirical findings rather than only a demonstration: projection
geometry is exact, conditional localization is spatially accurate, and cross-view correspondence, not geometry, is
the dominant barrier to end-to-end coverage. Its present application value lies in fracture highlighting, rib-level
organization, and high-precision committed localizations under a conservative confidence policy; the low end-to-end
commitment yield (2.5\%, with 6 of 55 cases producing any reconstruction) makes it low-coverage rather than a
comprehensive reconstructor, and it inherits the limitations above. The staged analysis also points
the way forward: because detector quality was the effective lever within the tested system, we would prioritize
lateral-detector quality over the correspondence algorithm, and we explicitly leave open the broader correspondence
design space, i.e., global anatomical matching, rib identity and ordering constraints, graph-based or full-view
feature correspondence, and jointly learned detection-correspondence models, along with real (non-simulated)
biplanar inputs and a 3D confirmation pass over accepted points. The lasting contribution is a reproducible,
sealed-cohort framework for confidence-gated selective 3D localization that identifies cross-view correspondence as
the operational bottleneck and lateral-detector quality as the primary lever for improving coverage. The paper's
principal point, however, is not maximizing automatic reconstruction coverage but demonstrating that confidence-aware
selective localization can provide reliable 3D guidance without requiring unsupported predictions for every detected
fracture, combining selective prediction, anatomical organization, and confidence-aware abstention into a clinically
plausible assistive framework for projection-based fracture review.

\section*{Code Availability}
Source code, split identifiers, frozen policies, evaluation scripts, and reproducibility instructions are
available in the RibAssist~3D repository~\cite{soboka2026ribassistcode}. Third-party datasets and trained
checkpoints are not redistributed because of licensing and artifact-size constraints.

\bibliographystyle{ACM-Reference-Format}
\bibliography{refs}

\end{document}